%% file: main.tex
\documentclass[10pt,twocolumn,letterpaper]{article}
\usepackage[pagenumbers]{iccv}
\usepackage[table,xcdraw, dvipsnames]{xcolor}

\definecolor{iccvblue}{rgb}{0.21,0.49,0.74}

\usepackage[pagebackref,breaklinks,colorlinks,allcolors=iccvblue]{hyperref}

\usepackage{soul}
\usepackage[margin=1in]{geometry}
\usepackage{multirow}
\usepackage{booktabs}

\usepackage{listings}
\usepackage{arydshln}
\definecolor{mygreen}{rgb}{0,0.6,0}   
\definecolor{myblue}{rgb}{0,0,0.8}      
\definecolor{myblack}{rgb}{0,0,0}       
\lstdefinestyle{mystyle}{
    backgroundcolor=\color{white},
    commentstyle=\color{mygreen}\ttfamily,
    keywordstyle=\color{myblue}\ttfamily,
    basicstyle=\footnotesize\ttfamily,
    breaklines=true,
    numbers=left,
    numbersep=5pt,
    frame=single,
    captionpos=b,
    xleftmargin=0pt,
    xrightmargin=0pt,
}
\title{Similarity-Aware Token Pruning: Your VLM but Faster}

\author{
    Ahmadreza Jeddi$^{1,2,4}$
    \and
    Negin Baghbanzadeh$^{2,3}$
    \and
    Elham Dolatabadi$^{2,3}$
    \and
    Babak Taati$^{1,2,4}$\\
    \normalsize
    $^1$University of Toronto \quad
    $^2$Vector Institute \quad
    $^3$York University, Canada \quad
    $^4$KITE Research Institute, UHN\\
    \normalsize
    \textbf{Corresponding author:} Ahmadreza Jeddi \texttt{(ahmadreza.jeddi@mail.utoronto.ca)}
}

\begin{document}
\maketitle

\input{sec/abstract}
\input{sec/intro}

\input{sec/related_work}

\input{sec/analysis}

\input{sec/method}

\input{sec/vit_results}

\input{sec/vlm_results}


\bibliographystyle{plain}
\bibliography{main} 

\clearpage
\setcounter{page}{1}
\maketitlesupplementary
\input{sec/supplementary}

\end{document}

%% file: sec/abstract.tex
\definecolor{customred}{HTML}{ED028C}  

\hypersetup{
    colorlinks=true,
    urlcolor=customred  
}

\begin{abstract}
The computational demands of Vision Transformers
(ViTs) and Vision-Language Models (VLMs) remain a
significant challenge due to the quadratic complexity
of self-attention. While token pruning offers a promising solution, existing methods often introduce training
overhead or fail to adapt dynamically across layers.
We present SAINT, a training-free token pruning framework that leverages token similarity and a graph-based
formulation to dynamically optimize pruning rates and
redundancy thresholds. Through systematic analysis,
we identify a universal three-stage token evolution process (aligner-explorer-aggregator) in transformers, enabling aggressive pruning in early stages without sacrificing critical information. For ViTs, SAINT doubles the
throughput of ViT-H/14 @224px with only 0.6\% accuracy loss on ImageNet-1K, surpassing the closest competitor by 0.8\%. For VLMs, we apply SAINT in three modes, ViT-only, LLM-only, and hybrid. SAINT reduces
LLaVA-13B’s tokens by 75\%, achieving latency comparable to LLaVA-7B with $< 1\%$ performance loss across benchmarks. Our work establishes a unified, practical framework for efficient inference in ViTs and VLMs.
Code available at: \href{https://github.com/ArmenJeddi/saint}{\textcolor{customred}{https://github.com/ArmenJeddi/saint}}
\end{abstract}

%% file: sec/intro.tex
\section{Introduction}
\label{sec:intro}
ViTs have become the standard in modern computer vision and are widely applied across tasks~\cite{dosovitskiy2020image}. In particular, VLMs that integrate large-scale ViTs with Large Language Models (LLMs) have gained significant traction~\cite{liu2023visual, wang2024qwen2, li2024mini, dai2023instructblip}. However, this impressive performance comes with a high computational cost due to the quadratic complexity of global self-attention~\cite{vaswani2017attention}. As models scale up and input resolutions increase, both training and inference become more challenging~\cite{li2023otterhd}.

A common strategy to reduce this cost is token pruning, which removes redundant tokens while preserving critical information. Many ViT-based methods integrate pruning into training or fine-tuning, adding modules to learn token importance or optimal drop ratios~\cite{kong2022spvit, yin2022vit, rao2021dynamicvit, chen2023diffrate}. Similarly, VLM approaches refine the vision backbone through token pooling or selection mechanisms~\cite{jaegle2021perceiver, li2024llama, alayrac2022flamingo}. While effective, these techniques introduce extra parameters and computational overhead, often requiring substantial compute. In contrast, training-free methods provide an attractive alternative, and in this work, we focus exclusively on inference-time token pruning.

Training-free token pruning techniques can be broadly categorized into two groups. \textit{First}, \textbf{the ranking strategy}. Some methods leverage the attention matrix~\cite{fayyaz2022adaptive, wang2024zero} or its approximations~\cite{yang2024visionzip, xing2024pyramiddrop} to assess token importance, while more recent works estimate redundancy through token similarities, often using the cosine similarity of key embeddings~\cite{bolya2022token, kim2024token, wu2023ppt}. \textit{Second}, \textbf{the removal strategy}. Some approaches drop low-importance tokens~\cite{fayyaz2022adaptive, chen2024image}, whereas others merge redundant tokens with their neighbors to preserve information~\cite{bolya2022token, yang2024visionzip, xu2024gtp}. While many such methods have been proposed for ViTs, the landscape for VLMs is sparser. Recent VLM token pruning efforts, such as FastV~\cite{chen2024image}, PyramidDrop~\cite{xing2024pyramiddrop}, VisionZip~\cite{yang2024visionzip} and SparseVLM~\cite{zhang2024sparsevlm}, rely primarily on attention matrix, with no prior work systematically exploring similarity-based pruning.

Despite notable progress, current methods still exhibit limitations. Many approaches focus primarily on reducing theoretical FLOPs without thoroughly analyzing how token dynamics evolve across transformer layers. Furthermore, most studies concentrate on a limited set of architectures, often overlooking intuitive insights, such as the benefits of pruning more tokens in early layers. Moreover, token pruning in ViTs and VLMs is typically studied separately, despite their shared challenges. This fragmented approach highlights the need for a unified framework to address token redundancy comprehensively.

Building on these limitations, we conduct an in-depth study of token evolution, analyzing not only the attention distributions of the CLS token but also token similarities across layers. Our findings suggest a general pattern across architectures: transformers tend to follow an \emph{aligner explorer aggregator} process, with significant token redundancy in the early (\emph{aligner}) and late (\emph{aggregator}) stages. Notably, in these stages, particularly the \emph{aligner}, a simple similarity aware dropping mechanism proves more effective than complex merging strategies. 

Leveraging these insights, we introduce Similarity Aware INference Time Token Pruning (SAINT), a lightweight, training free module that models tokens as graph nodes and dynamically determines both the optimal drop ratio and redundancy based on token similarity. By aggressively pruning tokens in early layers, SAINT achieves substantial efficiency gains without additional training or fine tuning, demonstrating broad applicability to both ViTs and VLMs.

We apply SAINT to both ViTs and VLMs with minimal modifications, achieving state-of-the-art accuracy-efficiency trade-offs. In ViTs, SAINT efficiently improves throughput while maintaining strong performance, surpassing methods that optimize FLOPs. For VLMs, we present the first study of training-free token pruning across three stages: text-agnostic (pruning before the language model), LLM-only (pruning within the LLM), and a hybrid approach. Through comprehensive ablation across all proposed configurations, we demonstrate that SAINT achieves competitive or superior performance across benchmarks.

In summary, our contributions are threefold: 
\begin{itemize} 
    \item We rigorously analyze token dynamics across transformer layers, identifying a universal three-stage process that guides efficient token pruning. 
    \item We introduce SAINT, a novel, training-free token pruning algorithm that leverages token similarities through a graph-based formulation to dynamically adjust pruning rates and redundancy ranking. 
    \item We demonstrate that SAINT achieves state-of-the-art accuracy-throughput trade-offs on ImageNet-1K~\cite{russakovsky2015imagenet} across ViT architectures and delivers substantial efficiency gains on VLMs, establishing a unified framework for inference-time token pruning. 
\end{itemize}

\begin{figure*}[t]
  \centering
  \includegraphics[width=\textwidth]{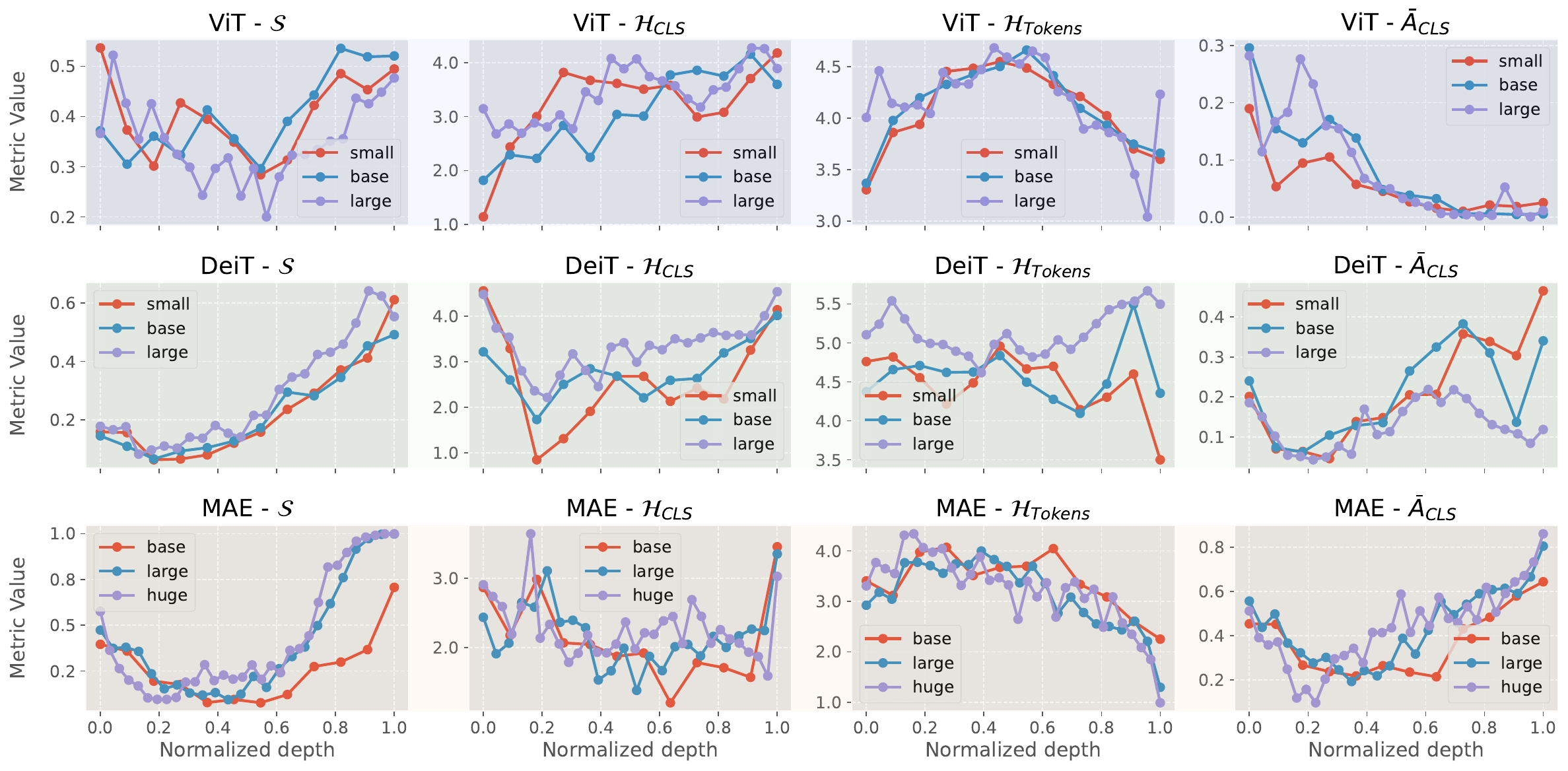}
  \caption{Token dynamics for different vision transformer models. The figure presents 12 panels arranged in a 3 (model class: ViT, i.e. trained on ImageNet, DeiT~\cite{touvron2021training}, MAE~\cite{he2022masked}) by 4 (metrics: $\mathcal{S}$, $\mathcal{H}_{CLS}$, $\mathcal{H}_{Tokens}$, $\bar{A}_{CLS}$) grid. Each panel shows the corresponding metric plotted against the normalized depth, with curves representing different model sizes.}
  \label{fig:token_dynamics}
\end{figure*}

%% file: sec/related_work.tex
\section{Related Work}
\label{sec:related_work}
\textbf{Token Pruning for ViTs:}
Several methods, such as SPViT~\cite{kong2022spvit}, EViT~\cite{liang2022not}, A-ViT~\cite{yin2022vit}, TPS~\cite{wei2023joint}, and DiffRate~\cite{chen2023diffrate}, require additional training or fine-tuning to learn effective pruning strategies. In contrast, training-free approaches have recently emerged as attractive alternatives. Notably, Token Merging (ToMe)~\cite{bolya2022token} was among the first to propose a training-free strategy by merging redundant tokens using a lightweight graph-partitioning approach. ToMe also established token key similarity as an indicator of redundancy, a concept that subsequent works such as ToFu~\cite{kim2024token}, PPT~\cite{wu2023ppt}, and Zero-TPrune~\cite{wang2024zero} have adopted. However, these methods typically rely on fixed pruning ratios and do not fully exploit the evolving token dynamics across transformer layers, missing opportunities for more aggressive pruning in early layers. SAINT addresses these limitations by leveraging a graph-based framework that employs a voting mechanism to adjust the pruning rate based on token similarity. This adaptive approach capitalizes on early-layer redundancy and generalizes across a wide range of ViT architectures.

\noindent\textbf{Token Pruning for VLMs:}  
VLMs such as LLaVa~\cite{liu2023visual, liu2024llavanext} process between 576 and several thousand visual tokens, with even higher counts for video inputs~\cite{chen2024image}. This large token volume, coupled with the sparse nature of visual data~\cite{man1982computational}, necessitates effective token compression. Both training-based and training-free methods have been explored. For example, architectures like LLaMa-Vid~\cite{li2024llama}, Flamingo~\cite{alayrac2022flamingo}, and Idefics~\cite{laurenccon2024matters} train VLMs by leveraging pruning techniques such as Perceiver Resampler~\cite{gong2023multimodal} during training. More recently, training-free, attention-based token removal methods, including FastV~\cite{chen2024image}, PyramidDrop~\cite{xing2024pyramiddrop}, SparseVLM~\cite{zhang2024sparsevlm}, and VisionZip~\cite{yang2024visionzip}, have been proposed to operate inside or before the language model. A common limitation of these approaches is their reliance on fixed drop rates or fixed pruning layers and their focus on only one component of the VLM pipeline, either the ViT or the LLM. SAINT extends similarity-based token dropping to VLMs via a dynamic framework, applicable either before the LLM (\emph{text-agnostic}), within it (\emph{LLM-only}), or both (\emph{hybrid}).

%% file: sec/analysis.tex
\section{Token Dynamics in Transformers}
\label{sec:analysis}
In this section, we investigate the evolution of token representations across various ViTs to uncover intrinsic properties governing token redundancy. Our objective is to establish a principled basis for token pruning by analyzing the impact of different pruning strategies, thereby motivating our proposed method, SAINT. To this end, we design two experiments. In the first, we track four metrics capturing attention patterns and token similarities across the network. In the second, we examine how distinct token pruning strategies affect performance at different stages, revealing that ViTs typically follow an implicit three-stage process during visual representation encoding. Our experiments indicate that these patterns also extend to pretrained CLIP~\cite{radford2021learning} and LLMs.

\subsection{Tracking Token Evolution}

To characterize token evolution, we monitor the following four metrics across the transformer depth.

\paragraph{Key Similarity.}\vspace{-5mm} Inspired by ToMe~\cite{bolya2022token}, we quantify token redundancy via the cosine similarity of head-averaged keys. For a ViT with $H$ attention heads:

\begin{equation}
\bar{k}_i = \frac{1}{H} \sum_{h=1}^{H} k_{h,i}, \quad i \in \{1,\dots,N\},
\end{equation}

\noindent where \( N \) is the number of tokens at a layer, and $k_{h,i}$ is key for token $i$ at the $h$-th head. The overall similarity score, denoted as \( \mathcal{S} \), is given by:

\begin{equation}
\mathcal{S} = \frac{1}{N^2} \sum_{i,j=1}^{N} \frac{\bar{k}_i \cdot \bar{k}_j}{\|\bar{k}_i\| \|\bar{k}_j\|}.
\end{equation}

\paragraph{CLS Token Attention Entropy.} \vspace{-10pt} This metric quantifies the entropy of the CLS token's attention distribution:
\begin{equation}
\mathcal{H}_{CLS} = -\sum_{j=1}^{N} \bar{A}_{1,j} \log \bar{A}_{1,j},
\end{equation}
where \( \bar{A}_{1,j} \) is the CLS token's head-averaged attention.

\paragraph{Average Token Attention Entropy.} \vspace{-10pt} To capture the dispersion of attention for non-CLS tokens, especially for models without CLS, we compute:
\begin{equation}
\mathcal{H}_{Tokens} = -\frac{1}{N-1}\sum_{i=2}^{N}\sum_{j=1}^{N} \bar{A}_{i,j} \log \bar{A}_{i,j}.
\end{equation}

\paragraph{Mean Normalized CLS Attention.} \vspace{-10pt} Finally, we assess token alignment with the global representation by measuring how tokens attend to CLS:
\begin{equation}
\bar{A}_{CLS} = \frac{1}{N}\sum_{i=1}^{N}\bar{A}_{i,1}.
\end{equation}

\begin{figure*}[t]
    \centering
    \includegraphics[width=\textwidth,height=0.4\textheight,keepaspectratio]{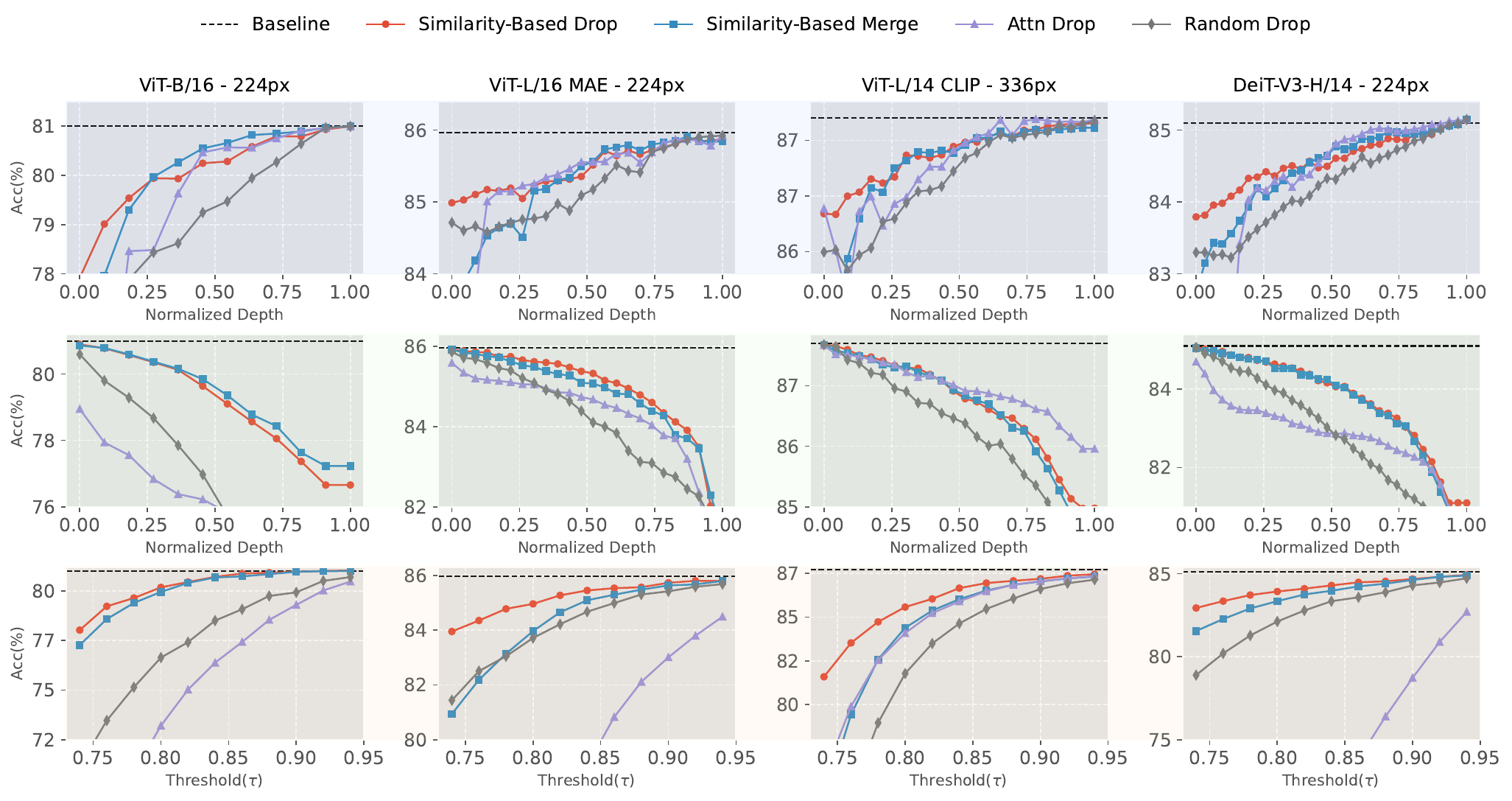}
    \vspace{-6mm}
    \caption{
        Impact of 4 pruning methods on model accuracy for 4 models on ImageNet. The subplots are arranged in 3 rows corresponding to different pruning strategies: (i) the top row prunes a large amount (r=40\%) from only one layer, (ii) the middle row prunes a constant per-layer amount (r = 100/depth\%) up to a specified layer, and (iii) the bottom row employs our voting-based method to determine the prune rate for the first half of the network layers. The baseline (no pruning) is included for reference.
    }

    \label{fig:pruning_analysis}
\end{figure*}

Collectively, these metrics provide a holistic view of token dynamics across layers. \autoref{fig:token_dynamics} shows their evolution across ViT architectures and model scales. While attention-based metrics (\(\mathcal{H}_{CLS}\), \(\mathcal{H}_{Tokens}\)) display higher variance across configurations, the key similarity metric \(\mathcal{S}\) demonstrates remarkably consistent patterns. This comparative stability suggests \(\mathcal{S}\) offers a more reliable basis for redundancy detection than attention distributions alone, particularly when generalizing across architectures.

Our experiments further hint at an implicit three-phase progression in ViTs that loosely aligns with what we term the \emph{aligner-explorer-aggregator} paradigm. In the early \emph{aligner} stage, tokens appear to align with one another and with the global CLS token, as indicated by elevated \(\mathcal{S}\) and \(\bar{A}_{CLS}\). An intermediate phase is then observed where increased token diversity corresponds to a decrease in \(\mathcal{S}\) and an increase in \(\mathcal{H}_{Tokens}\), suggesting a broadening of attention patterns. Finally, a later phase seems to involve a reconvergence into a more homogeneous representation, evidenced by a resurgence in \(\mathcal{S}\) and \(\bar{A}_{CLS}\) alongside lower \(\mathcal{H}_{Tokens}\). Although prior works have hinted at similar phases~\cite{park2023self, clark2019does}, our results are broadly consistent with this progression, while also indicating some variations across different architectures and training conditions.

Motivated by the stage specific dynamics revealed in \autoref{fig:token_dynamics}, we compare attention-based versus similarity-based scoring mechanisms, as well as token dropping versus merging strategies, across the identified \emph{aligner}, \emph{explorer}, and \emph{aggregator} stages.

\begin{figure*}[t]
    \centering
    \includegraphics[width=\textwidth,height=0.28\textheight,keepaspectratio]{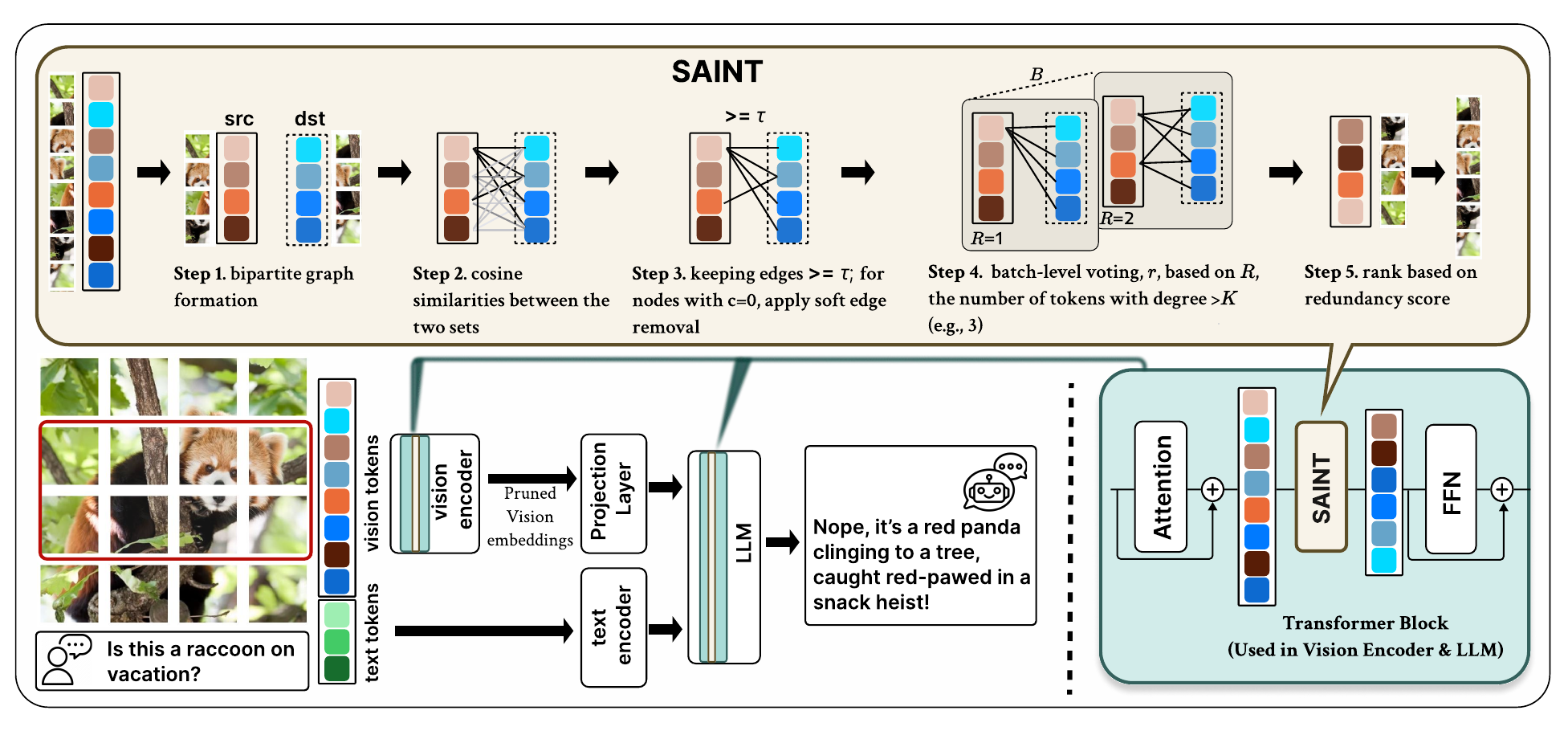}
    \caption{Overview of SAINT: Inserted between attention and feed-forward blocks, SAINT models tokens as nodes in a bipartite graph with edges weighted by cosine similarity. It prunes redundant tokens via thresholding, batch-level voting, and redundancy-based ranking. SAINT integrates into transformer layers in vision encoders, LLMs, or both.}

    \label{fig:saint_overview}
\end{figure*}

\subsection{Analysis of Pruning Strategies}
To evaluate how different pruning strategies affect network performance across the three stages, we experiment with four straightforward methods: (i) attention-based dropping, where tokens are ranked by CLS attention, (ii) similarity-based merging, akin to ToMe, (iii) similarity-based dropping, which removes instead of merges tokens, and (iv) random dropping as a baseline. Our evaluation is conducted in two settings.

\noindent\textbf{Single Layer Pruning Experiments.} Here, a fixed high percentage of tokens is pruned from a single layer. The first row of \autoref{fig:pruning_analysis} shows that similarity-based token dropping minimizes accuracy loss most effectively in the \emph{aligner} stage, making early pruning more viable and yielding better computational gains. \textbf{Observation: Similarity-aware token dropping is most robust at the early stage compared to other methods.} In the \emph{explorer} and \emph{aggregator} stages, iterative attention pooling distributes token information, resulting in comparable performance across methods.

\noindent\textbf{Progressive Pruning Experiments.} Here, tokens are pruned from the input layer to a fixed depth. \autoref{fig:pruning_analysis} row 2 shows results for a constant drop ratio per layer, while row 3 adopts an adaptive approach: pruning is confined to the first half of the transformer layers, with drop rates determined by our adaptive voting-based method (Eq.~\eqref{adaptive_prune_rate}) and a token similarity threshold, \(\tau\).

The setup in row 2 is common among recent methods like ToMe and ToFu. Even in this scenario, similarity-based approaches outperform attention-based pruning. However, as pruning extends deeper into the network, performance degradation accelerates. \textbf{Observation: Pruning at later stages significantly harms performance without substantial gains.} While late-stage pruning may reduce FLOPs, its efficiency gains are minimal compared to early-stage pruning, making it a less favorable trade-off. This limitation is often overlooked.

Given the robustness of similarity-based token dropping in the \emph{aligner} stage (\autoref{fig:pruning_analysis}, row 1) and the diminishing returns of late-stage pruning (\autoref{fig:pruning_analysis}, row 2), we propose SAINT. Our method leverages token redundancies measured via key similarity to maximize early-layer pruning while reducing performance loss (\autoref{fig:pruning_analysis}, row 3), optimizing for throughput and latency over FLOPs.

%% file: sec/method.tex
\section{Similarity-Aware INference Time Token Dropping (SAINT)}
\label{sec:saint}

We present SAINT, a plug and play module integrated into transformer layers (\autoref{fig:saint_overview}). Leveraging graph theoretic principles, SAINT ranks tokens based on redundancy measured by key similarity and drops the least informative ones during inference. Our approach is motivated by Section~\ref{sec:analysis}, which shows that key similarity robustly indicates redundancy and that early layer token pruning offers significant efficiency gains.

\subsection{Graph Construction and Edge Selection}

At a transformer layer \(L\), let token representations be \(X \in \mathbb{R}^{B \times N \times C}\) and head-averaged keys \(\bar{K} \in \mathbb{R}^{B \times N \times \frac{C}{H}}\), where \(B\) is batch size, \(N\) tokens, \(C\) channels, and \(H\) heads. We view each token as a node \(v_i \in V\) in graph \(G=(V,E)\) and partition them into two disjoint sets (src and dst) using an alternating scheme, forming the bipartite graph \(G_{\text{bip}}=(V_{\text{src}}, V_{\text{dst}}, E)\). We then compute the cosine similarity between tokens in these sets to obtain the weighted adjacency matrix:
\[
S_{ij} = \frac{\bar{k}^{\text{src}}_i \cdot \bar{k}^{\text{dst}}_j}{\|\bar{k}^{\text{src}}_i\| \|\bar{k}^{\text{dst}}_j\|}.
\]

Instead of retaining only the maximum similarity per token~\cite{bolya2022token, kim2024token}, we retain all edges with \(S_{ij} \ge \tau\) (with \(\tau\) a user-defined threshold), thereby defining a subgraph \(G_\tau\) that captures each token's local neighborhood. The degree of a node \(v_i\) is then defined as:
\begin{equation}
d(v_i) = \sum_{j \in V_{\text{dst}}} \mathbb{I}\{S_{ij} \ge \tau\},
\end{equation}
which serves as a natural measure of redundancy: tokens with high degree connect to many similar neighbors.

\begin{figure*}[t]
    \centering
    \includegraphics[width=\linewidth]{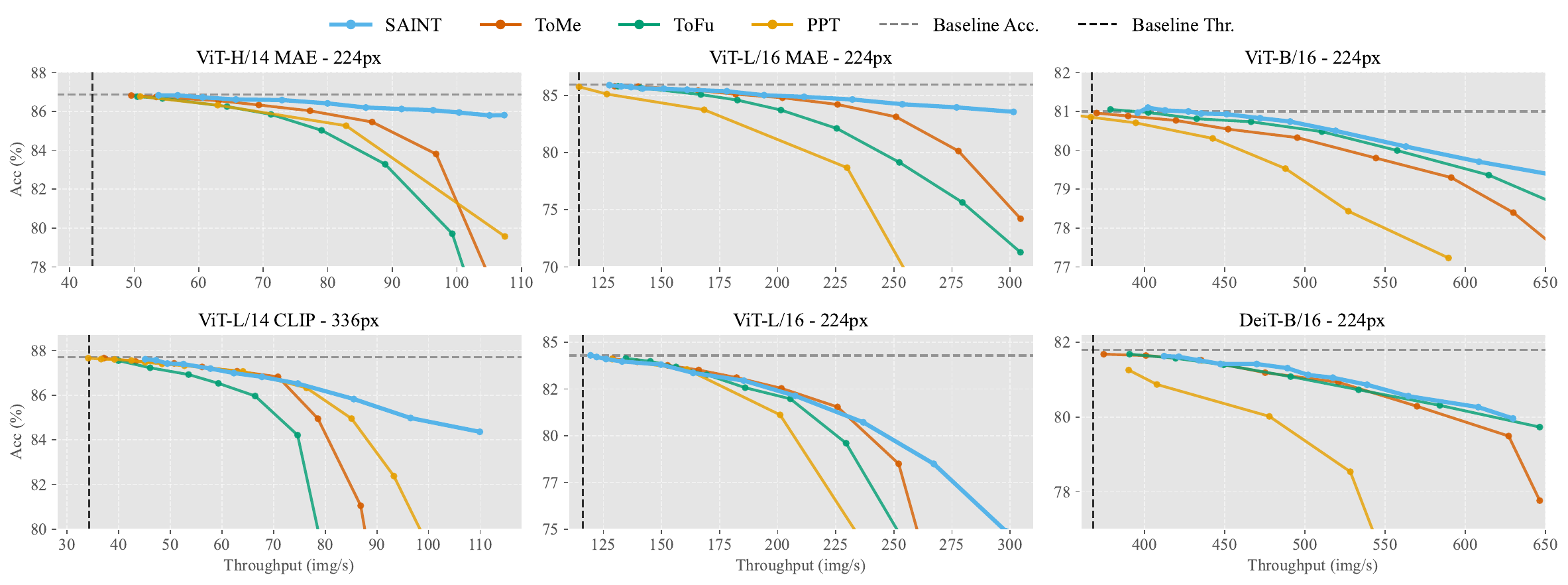}
    \vspace{-7mm}
    \caption{Accuracy/Throughput trade-off for SAINT and baseline methods across various ViT paradigms}
    \label{fig:model_sweep}
\end{figure*}

\subsection{Adaptive Prune Rate via Voting}

To maximize efficiency gains, particularly in early layers, our goal is to drop as many redundant tokens as possible without compromising performance. Unlike methods that enforce a fixed drop rate to maintain a uniform token count~\cite{kim2024token, bolya2022token, wu2023ppt}, SAINT employs a voting mechanism based on the node degrees in \(G_\tau\). 

We introduce a hyperparameter \(K\) (analogous to the number of neighbors in a KNN graph) such that a token is considered highly redundant if \(d(v_i) \ge K\). For each sample in a batch, let \(R\) be the number of tokens satisfying this condition. The adaptive prune rate \(r\) for the layer is determined by batch-level voting:
\begin{equation} \label{adaptive_prune_rate}
r = \left\lfloor \frac{1}{B} \sum_{b=1}^{B} R^{(b)} \right\rfloor.
\end{equation}
This formulation leverages local connectivity to adjust token drop rates dynamically, enabling more aggressive pruning in early layers with high redundancy

\subsection{Redundancy Ranking and Token Dropping}

After determining the adaptive prune rate \(r\), we rank tokens based on a redundancy score that incorporates both their connectivity and the quality of their connections. For each token \(v_i \in V_{\text{src}}\), we compute:
\begin{enumerate}
    \item The degree \(d(v_i)\).
    \item The mean similarity over its valid edges:
    \begin{equation}
    m(v_i) = \frac{1}{d(v_i)} \sum_{j \in V_{\text{dst}}} S_{ij} \cdot \mathbb{I}\{S_{ij} \ge \tau\}.
    \end{equation}
\end{enumerate}
We then define the redundancy score as:
\begin{equation}\label{rank_eq}
\text{score}(v_i) =
\begin{cases}
d(v_i) \cdot e^{\gamma (m(v_i) - \tau)}, & \text{if } d(v_i) > 0, \\
\tilde{m}(v_i), & \text{otherwise},
\end{cases}
\end{equation}
where \(\gamma\) is a scaling factor and \(\tilde{m}(v_i)\) is an alternative score for tokens with no valid edges. This score can be interpreted as a \textbf{centrality} measure: tokens with high degree and high average similarity are considered redundant. Tokens are then ranked by \(\text{score}(v_i)\), and the \(r\) tokens with the highest scores are dropped.

In summary, token pruning is cast as a graph sparsification problem where tokens are nodes, cosine similarities form weighted edges, and redundancy is measured by node degree and centrality. SAINT reduces inference load while preserving key tokens. For more details, see Supp~1.4 for a PyTorch implementation.

%% file: sec/vit_results.tex
\section{ViT Experiments}
\label{sec:vit_experiments}
We first analyze SAINT's design through ablation studies, then evaluate our optimal configuration on ImageNet-1K~\cite{russakovsky2015imagenet} using diverse ViTs, including supervised (DeiT~\cite{touvron2021training}) and self-supervised variants (MAE~\cite{he2022masked}, CLIP~\cite{radford2021learning}). Results for VLMs follow in Section~\ref{sec:vlm_experiments}. Though SAINT supports both training and inference phases, we focus exclusively on inference-time efficiency. Throughput is measured on an A40 GPU using optimal batch sizes in FP32, with mean accuracy/throughput reported over five runs to account for batch statistic variance. While SAINT's key-similarity approach inherently avoids attention matrix computation, making it compatible with FlashAttention~\cite{dao2205fast,dao2023flashattention} and PyTorch SDPA, we maintain consistency by evaluating all methods with standard eager-mode attention. Further implementation details for the ViT case are provided in Supp.~1.1.

\newcommand{\highlight}[2]{\colorbox[HTML]{#1}{\color{black}#2}}

\begin{table}[t]
    \centering
    \caption{Ablation of design choices for SAINT ($\tau=0.75$ in all experiments) on a ViT-L/16 MAE (224px) model (baseline: 85.96\% accuracy, 115.6 img/s throughput). Rows highlighted in \highlight{96FFFB}{\kern-\fboxsep blue\kern-\fboxsep} indicate our default SAINT setup, which for this model more than doubles throughput with $\sim 1\%$ accuracy drop.}
    \label{tab:ablation_vit}
    \footnotesize
    \setlength{\tabcolsep}{6pt}
    \begin{tabular}{llcc}
    \toprule
    \textbf{Metric} & \textbf{Option} & \textbf{Acc (\%)} & \textbf{img/s} \\
    \midrule
    \multirow{2}{*}{Graph} 
        & Full            & 82.83                         & 301.5                        \\
        & BiPartite       & \cellcolor[HTML]{96FFFB}84.74 & \cellcolor[HTML]{96FFFB}238.6 \\
    \midrule
    \multirow{2}{*}{Drop Rate}
        & Constant (r=10) & 84.65                         & 214.1                         \\
        & K-voting        & \cellcolor[HTML]{96FFFB}84.74 & \cellcolor[HTML]{96FFFB}238.6 \\
    \midrule
    \multirow{4}{*}{K}
        & 1               & 79.62                         & 500.3                         \\
        & 3               & 84.03                         & 281.4                         \\
        & 5               & \cellcolor[HTML]{96FFFB}84.74 & \cellcolor[HTML]{96FFFB}238.6 \\
        & 10              & 85.32                         & 180.9                        \\
    \midrule
    \multirow{2}{*}{Ranking}
        & Eq. \eqref{rank_eq} & \cellcolor[HTML]{96FFFB}84.74 & \cellcolor[HTML]{96FFFB}238.6 \\
        & \(m(v_i)\)                      & 84.68                         & 234.3                         \\
    \midrule
    \multirow{3}{*}{\(\gamma\)}
        & 1               & 84.73                         & 236.4                         \\
        & 10              & \cellcolor[HTML]{96FFFB}84.74 & \cellcolor[HTML]{96FFFB}238.6 \\
        & 100             & 84.79                         & 234.3                         \\
    \bottomrule
    \end{tabular}
\end{table}

\subsection{Design Choices for SAINT}
\label{subsec:vit_design}

\autoref{tab:ablation_vit} presents an ablation study on a ViT-L/16 (MAE) model, examining key design choices in SAINT. We compare bipartite versus full graph partitioning, the neighbor threshold \(K\), the scaling parameter \(\gamma\), and our voting-based drop rate and ranking algorithm against baselines. Our default configuration (highlighted in blue) achieves a strong balance between accuracy and throughput, with \(K=5\) and \(\gamma=10\) consistently yielding robust results. Additionally, our dynamic drop rate calculation improves the trade-off, while the ranking algorithm enhances robustness.

In \autoref{tab:ablation_vit}, we use a fixed \(\tau=0.75\) and apply SAINT in the first half of the network. As discussed in Section~\ref{sec:analysis}, these parameters are crucial for optimizing the performance-efficiency trade-off. Ablation studies to investigate their impact are presented in Supp~2.1 and 2.2.

\subsection{Model Sweep}
\label{subsec:model_sweep}

We evaluate SAINT's accuracy-throughput trade-off across six ViT models. \autoref{fig:model_sweep} compares SAINT against PPT~\cite{wu2023ppt}, ToMe~\cite{bolya2022token}, and ToFu~\cite{kim2024token}. Across all models, SAINT consistently achieves a favorable trade-off, offering significant throughput gains with minimal accuracy loss. Interestingly, even within the same architecture (ViT-L/16), pruning behavior varies with the training regimen. SAINT proves particularly robust at higher throughput gains, largely due to its early-stage token removal strategy. As shown in ~\autoref{fig:model_sweep}, SAINT performs on par or outperforms all competitors across different ViT variants, resolutions, and pruning rates. For example, on a ViT-H/14 (MAE), at $2\times$ the baseline throughput, SAINT loses only $~0.6\%$ accuracy, which is $~0.8\%$ higher than the closest competitor (ToMe).

In Supp~3, we compare SAINT to other methods in FLOPs. While SAINT does not achieve the lowest FLOPs, it delivers superior accuracy and inference speed, showing that FLOPs alone do not dictate real-world efficiency. This highlights that optimizing purely for FLOPs does not always improve inference latency, reinforcing SAINT's practical advantage.

%% file: sec/vlm_results.tex
\newcommand{\negin}[1]{{\color{orange}[\hl{NB}: #1]}}

\section{VLM Experiments}
\label{sec:vlm_experiments}

We apply SAINT to VLMs to enhance their inference efficiency. Prior training-free methods have focused on attention-based pruning, typically using either text-agnostic approaches~\cite{yang2024visionzip} or strategies limited to LLM layers~\cite{chen2024image,xing2024pyramiddrop,zhang2024sparsevlm}. Using LLaVA models~\cite{liu2023visual} as our backbone, we evaluate SAINT against existing baselines across multiple visual understanding benchmarks. Additionally, we analyze VLM performance in text-agnostic, LLM-only, and hybrid configurations.

For evaluation, we follow standard VLM protocols, testing single-sample inference (batch size=1) with LLaVA 7B and 13B variants. Our analysis covers core tasks like image captioning and visual question answering (VQA), using benchmarks such as POPE~\cite{Li-hallucination-2023} for hallucination analysis and MME~\cite{fu2023mme} for multimodal understanding assessment. Our experiments measure both accuracy preservation and latency reduction. For more details on VLM implementation details and evaulation benchmarks refer to Supp 1.2, 1.3.

\begin{table}[t]
    \centering
    \footnotesize 
    \renewcommand{\arraystretch}{0.85} 
    \setlength{\tabcolsep}{3pt} 

    \caption{Comparison of different models using LLM-Only pruning strategies across various benchmarks on LLaVA-7B, evaluated at different token retention sizes. The best performance is highlighted in bold, the second-best is underlined.}
    \label{tab:LLM-Only-Comparison}
    
    \resizebox{\columnwidth}{!}{%
        \begin{tabular}{lcccccc}
            \toprule
            \multicolumn{1}{c}{} & \multicolumn{6}{c}{\textbf{Baseline - 576 tokens}} \\
            \cmidrule(lr){2-7}
            Model & MME & POPE & GQA & VizWiz & MMB$^{\text{cn}}$ & SQA$^{\text{I}}$ \\
            \midrule
            LLaVA-1.5-7B & 1511 & 86.97 & 61.93 & 54.3 & 55.67 & 69.46 \\
            \midrule
            \multicolumn{1}{c}{} & \multicolumn{6}{c}{\textbf{Retain Average 288 tokens}} \\
            \cmidrule(lr){2-7}
            FastV & 1460 & 82.9 & 57.8 & 50.2 & 52.6 & 67.2 \\
            SparseVLM & 1472 & 85.9 & \underline{61.1} & 53.8 & 55.3 & 68.9 \\
            PyramidDrop & \underline{1505} & \underline{86.1} & 60.9 & \textbf{54.1} & \underline{55.6} & \underline{69.3} \\
            SAINT (Ours) & \textbf{1510} & \textbf{86.9} & \textbf{61.5} & \underline{54.01} & \textbf{55.9} & \textbf{69.7} \\
            \midrule
            \multicolumn{1}{c}{} & \multicolumn{6}{c}{\textbf{Retain Average 144 tokens}} \\
            \cmidrule(lr){2-7}
            FastV & 1417 & 79.1 & 54.2 & 46.1 & 47.2 & 60.1 \\
            SparseVLM & 1451 & 85.2 & 59.9 & 53.4 & 54.6 & 68.4 \\
            PyramidDrop & \underline{1499} & \underline{86.1} & \underline{60.7} & \underline{53.6} & \textbf{55.2} & \underline{69.2} \\
            SAINT (Ours) & \textbf{1500} & \textbf{86.5} & \textbf{61.2} & \textbf{53.8} & \underline{54.8} & \textbf{69.6} \\
            \midrule
            \multicolumn{1}{c}{} & \multicolumn{6}{c}{\textbf{Retain Average 72 tokens}} \\
            \cmidrule(lr){2-7}
            FastV & 1394 & 72.3 & 50.3 & 40.2 & 43.7 & 53.4 \\
            SparseVLM & 1425 & 84.5 & \underline{60.5} & 52.1 & 54.1 & 66.2 \\
            PyramidDrop & \underline{1480} & \textbf{86.1} & 60.1 & \underline{53.4} & \textbf{55.0} & \textbf{69.2} \\
            SAINT (Ours) & \textbf{1492} & \underline{85.9} & \textbf{60.7} & \textbf{53.5} & \underline{54.6} & \underline{68.9} \\
            \bottomrule
        \end{tabular}%
    }
    
\end{table}

\begin{table}[t]
\centering
\renewcommand{\arraystretch}{1.1} 
\setlength{\tabcolsep}{4pt} 
\small 
\caption{Performance and latency comparison of LLaVA-7B, LLaVA-13B, and LLaVA-13B (+SAINT) across benchmarks. Best performance is bold, second-best underlined.}
\label{tab:llava_comparison}
\resizebox{\columnwidth}{!}{ 
\begin{tabular}{lccccccc}
\toprule
\textbf{LLaVA-1.5 Ver.} & \textbf{Avg. Latency} & \textbf{MME} & \textbf{POPE} & \textbf{GQA} & \textbf{SQA$^{\text{I}}$} & \textbf{MMB$^{\text{cn}}$} & \textbf{VizWiz} \\
\midrule
7B  & \textbf{244.1} & 1511 & 86.9 & 61.9 & 69.4 & 55.6 & 54.3 \\
13B & 379.3 & \textbf{1530} & \underline{87.14} & \textbf{63.3} &\textbf{ 72.8} & \textbf{62.2} & \underline{56.5} \\
13B (+ SAINT)  & \underline{268.1} & \underline{1523} & \textbf{87.2} & \underline{63.2} & \underline{72.7} & \underline{61.8} & \textbf{57.0} \\
\bottomrule
\end{tabular}
}
\end{table}

\subsection{Text-Agnostic (Pre-LLM) Pruning}
Token pruning is applied to the visual tokens generated by the ViT before they reach the LLM. Instead of iteratively pruning tokens across multiple ViT layers, we use a slightly modified variant of SAINT that applies similarity-aware pruning before the VLM projection layer, specifically after the 23\textsuperscript{rd} layer of CLIP in LLaVA, minimizing information loss. More details on this variant are provided in the Supp.~2.3.

\begin{figure}[t] 
    \centering
    \includegraphics[width=\linewidth]{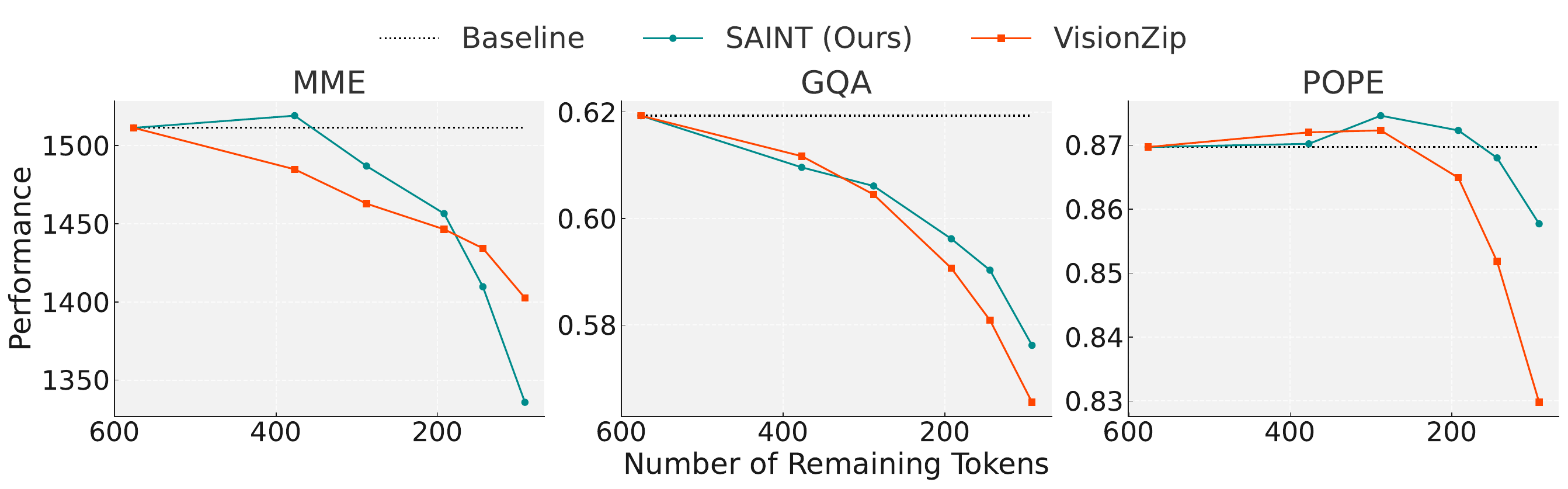} 
    \caption{The effect of text-agnostic pruning versus VisionZip on performance across three benchmarks using the LLaVa-1.5-7B model, plotted against the number of remaining tokens.}
    \label{fig:pre_llm}
\end{figure}

This approach is simple and efficient, reducing context size and compressing the KV cache at every LLM layer. \autoref{fig:pre_llm} examines text-agnostic pruning at different token scales on three benchmarks using LLaVA-7B, comparing SAINT to VisionZip~\cite{yang2024visionzip}. SAINT generally achieves higher scores, showing that its similarity-aware pruning better preserves critical visual details.

A key limitation, however, is its text-agnostic nature. In cases of severe pruning, important visual tokens relevant to specific user queries may be lost, as suggested by the results in \autoref{fig:pre_llm} and noted in~\cite{zhang2024sparsevlm}. Despite this, SAINT’s strong performance across diverse benchmarks highlights similarity-aware pruning as an effective pre-LLM strategy for improving inference efficiency.

\begin{table*}[t]
\centering
\renewcommand{\arraystretch}{1.2} 
\setlength{\tabcolsep}{6pt} 
\footnotesize 
\caption{Performance and latency comparison of SAINT approaches with 144 average remaining tokens. Best results are in bold, second-best underlined. Latency represents the average output generation time per sample in ms.}
\label{tab:saint_comparison}
\resizebox{\textwidth}{!}{ 
\begin{tabular}{lcccccc cccccc c}
\toprule
\textbf{Model} & \multicolumn{6}{c}{\textbf{Performance}} & \multicolumn{6}{c}{\textbf{Latency (ms)}} & \textbf{Avg. Latency} \\
\cmidrule(lr){2-7} \cmidrule(lr){8-13}
 & \textbf{MME}  & \textbf{POPE} & \textbf{GQA} & \textbf{SQA$^{\text{I}}$} & \textbf{MMB$^{\text{cn}}$} & \textbf{VizWiz} & \textbf{MME}  & \textbf{POPE}  & \textbf{GQA} & \textbf{SQA$^{\text{I}}$} & \textbf{MMB$^{\text{cn}}$} & \textbf{VizWiz} & \\
\midrule
ViT-Only  & 1409 & \underline{86.6} & 59.0 & 68.2 & 52.7 & \textbf{54.7} & \textbf{129} & 166 & \textbf{142} & \textbf{140} & \underline{181} & \underline{216} & \textbf{162.3} \\
LLM-Only  & \underline{1500} & 86.5 & \textbf{61.2} & \textbf{69.6} & \textbf{54.8} & 53.8 & 162 & \underline{153} & 173 & 169 & 189 & 229 & 179.2 \\
Hybrid       & \textbf{1506} & \textbf{86.8} & \underline{60.2} & \underline{68.6} & \underline{54.6} & \underline{53.9} & \underline{146} & \textbf{147} & \underline{159} & \underline{154} & \textbf{173} & \textbf{213} & \underline{165.3} \\
\bottomrule
\end{tabular}
}
\end{table*}

\subsection{LLM-Only Pruning}
SAINT naturally extends to language model transformers, where we observe pruning patterns similar to ViTs. Early-layer pruning enhances efficiency with mild performance degradation, motivating our default configuration: layers 8--16 with $K=5$, $\gamma=10$, and adaptive $\tau$ (layer selection ablated in Supp~2.3). Unlike attention-based alternatives, SAINT operates exclusively during prefilling stage, using visual token key similarities, enabling a compressed KV cache that accelerates subsequent decoding stages.

\autoref{tab:LLM-Only-Comparison} benchmarks SAINT against FastV~\cite{chen2024image}, PyramidDrop~\cite{xing2024pyramiddrop}, and SparseVLM~\cite{zhang2024sparsevlm} in the LLM-only configuration across varying token retention levels. While all baselines rely on attention-based ranking, their strategies differ: FastV prioritizes efficiency through aggressive early pruning (at layer 2), PyramidDrop follows a fixed layer-wise schedule (layers 8, 16, 24), and SparseVLM employs dynamic pruning rates. SAINT outperforms all three with similarity-aware adaptation. As results show, SAINT consistently matches or exceeds baseline performance. Notably, SAINT removes tokens earlier in transformer than PyramidDrop, improving both accuracy and latency.

\begin{figure}[t]
    \centering
    \includegraphics[width=\linewidth]{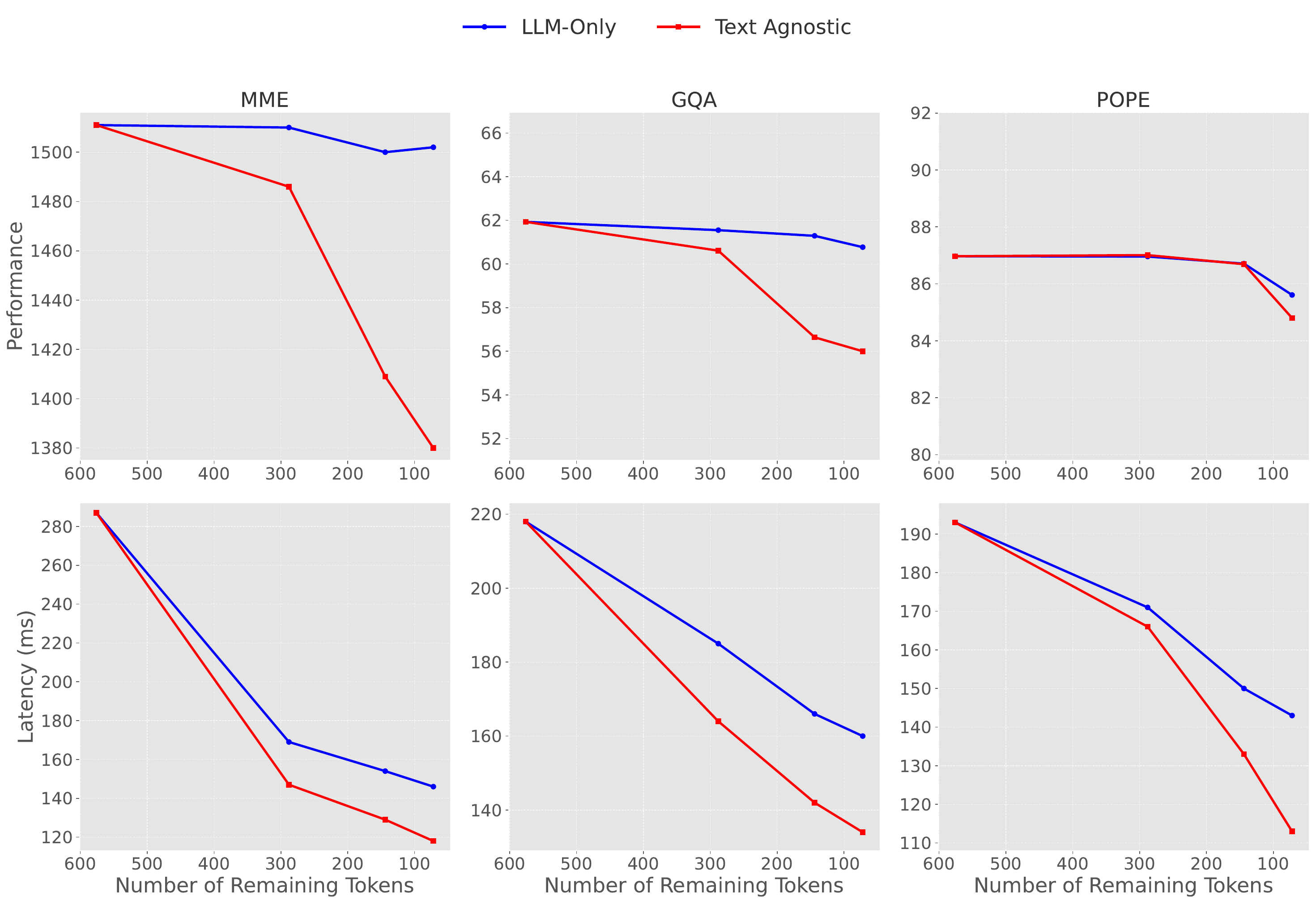}    
    \caption{Comparison of SAINT in the text-agnostic and LLM-only setups across three benchmarks. The top row presents the performance comparison, and the bottom row compares the latency of the two approaches.}
    \label{fig:performance_vs_generation_time}
\end{figure}

\autoref{fig:performance_vs_generation_time} further compares text-agnostic (ViT-only) and LLM-only pruning. While ViT-only pruning achieves superior efficiency, LLM-only pruning better preserves performance, highlighting the benefits of context-aware token reduction, especially at very low token rates.

A particularly notable experiment applies SAINT to LLaVA-13B, dropping 75\% of tokens. This reduces latency to nearly match LLaVA-7B while incurring less than a 1\% performance loss (\autoref{tab:llava_comparison}). This result demonstrates SAINT’s ability to improve larger models' practical viability without compromising accuracy. Additional results for LLaVA-13B are provided in Supp~2.3.

\subsection{Hybrid VLM Pruning}
Building on the efficiency-performance trade-offs observed in the previous sections, we explore a hybrid approach that combines both strategies. Specifically, we prune a moderate portion (~30\%) of tokens before the LLM while applying additional pruning within the LLM itself. This configuration aims to balance efficiency and performance.  

\autoref{tab:saint_comparison} compares the hybrid setup to ViT-only and LLM-only pruning, evaluating the trade-off between latency and performance. The results, shown for a total of 144 retained tokens, demonstrate how the hybrid approach benefits from the strengths of both techniques. Further exploration of this setup could yield even more favorable results.

\section{Conclusions}

Our empirical evaluation across both ViT and VLM settings demonstrates that SAINT achieves state-of-the-art efficiency gains while maintaining competitive performance. The throughput/accuracy improvements on ImageNet and the significant latency reductions on VLM tasks validate our approach and underscore its practical relevance in real-world applications. Detailed numerical results, additional ablations, and further discussion are provided in the Supplementary Material.

%% file: sec/supplementary.tex

\newcommand{\beginsupplement}{%
    \setcounter{section}{0}%
    \renewcommand{\thesection}{Supp \arabic{section}}%
}

\beginsupplement

\section{Implementation Details}
\label{appendix:implementation}

In this appendix, we provide additional details about our experimental setups and the PyTorch-style pseudocode for our SAINT module.

\subsection{Implementation Details for ViT Experiments on ImageNet-1K}
\label{appendix:implementation:vit}

For our ViT experiments on ImageNet-1K, we use the MAE checkpoints released by He \emph{et al.}~\cite{he2022masked} and load other ViT models (e.g., DINO, CLIP, DeiT, Supervised ViT) from the HuggingFace Timm library via its \texttt{create\_model} function. We follow the standard ImageNet-1K preprocessing pipeline (resizing, center-cropping to 224px, and normalizing) to prepare the inputs. Our SAINT module is inserted between each transformer block and its subsequent feed-forward network (FFN).

For baselines, we use the official Token Merging (ToMe) code without modifications. We implement Token Fusion (ToFu) ourselves based on their paper, and for PPT, we adapt their original DeiT-specific implementation to other ViT architectures.

\subsection{Implementation Details for VLMs}
\label{appendix:implementation:vlm}

We conduct experiments on LLaVa~1.5 with 7B and 13B parameters, using a ViT-L/CLIP backbone at an input resolution of 336px and a patch size of 14. This configuration yields $576$ patch tokens, plus a single \texttt{[CLS]} token. From the 23rd layer onward, the vision tokens are projected into the LLM input space, which corresponds to a Vicuna model with 32 transformer layers. We perform token pruning during the \emph{pre-fill} stage, before the text decoding begins.

We evaluate our VLM setups on multiple benchmarks, including MME, POPE, and GQA. In all cases, we insert SAINT into the ViT portion of the pipeline, although it can also be applied in a hybrid fashion within the language model (see Section~4.2 in the main paper).

\subsection{Evaluation Benchmarks for VLMs}
\label{appendix:evaluationbenchmarks:vlm}

In this section, we briefly describe the evaluation benchmarks used to assess model performance.\\
\textbf{VizWiz.} The VizWiz dataset~\cite{gurari2018vizwiz} consists of over 31,000 visual questions collected from blind individuals. Each image is captured using a mobile phone and is accompanied by a spoken question, later transcribed into text. The dataset presents unique challenges, including lower-quality images, conversational question formats, and unanswerable questions due to ambiguous content.\\
\textbf{MMBench.} MMBench~\cite{liu2024mmbench} provides a hierarchical evaluation of models through three levels of abilities: L-1 (perception and reasoning), L-2 (six sub-abilities), and L-3 (20 fine-grained dimensions). This structured framework enables a comprehensive analysis of multimodal understanding and reasoning.\\
\textbf{ScienceQA.} ScienceQA~\cite{lu2022learn} spans multiple domains, including natural sciences, language, and social sciences. It organizes questions into 26 topics, 127 categories, and 379 distinct skills, making it a robust benchmark for evaluating multimodal comprehension, multi-step reasoning, and interpretability.\\
\textbf{GQA.} The GQA benchmark~\cite{hudson2019gqa} tests visual reasoning and scene understanding using structured scene graphs. It includes questions designed to assess spatial relationships, object attributes, and logical inference, providing insights into a model’s ability to comprehend complex visual contexts.\\
\textbf{POPE.} The POPE dataset~\cite{li2023evaluating} evaluates object hallucination in models using binary questions about object presence in images. Accuracy, Recall, Precision, and F1 Score are used to quantify hallucination levels across different sampling strategies, ensuring a fine-grained analysis of model reliability.\\
\textbf{MME.} The MME benchmark~\cite{fu2023mme} assesses model performance across 14 subtasks targeting both perceptual and cognitive abilities. By employing manually curated instruction-answer pairs, MME mitigates data leakage, ensuring a fair and unbiased evaluation of multimodal learning capabilities.

\subsection{PyTorch Pseudocode for SAINT}
\label{appendix:implementation:saint}

Algorithm~1 presents the PyTorch-style pseudocode for our bipartite soft matching module, which constitutes the core of SAINT. The function \texttt{saint\_drop} returns a callable \texttt{drop\_func} that prunes a batch of tokens based on cosine similarity. We treat every pair of tokens (src/dst) as a bipartite match, compute their similarity, and drop those that are deemed redundant according to a threshold-based voting mechanism.

Algorithm~1 provides a self-contained PyTorch-style pseudocode snippet for our bipartite partitioning module. We treat tokens in pairs (src/dst), compute their cosine similarity, and then drop those that are deemed redundant according to a threshold-based voting mechanism.

\clearpage
\begin{center} \label{alg:saint}
\begin{minipage}{\textwidth}
\noindent\rule{\textwidth}{0.4pt}\\[1mm]
\centering\textbf{Algorithm 1: PyTorch implementation of SAINT}\\[1mm]
\noindent\rule{\textwidth}{0.4pt}\\[2mm]
\begin{lstlisting}[language=Python, frame=none, numbers=none, xleftmargin=0pt, xrightmargin=0pt]
# Inputs:
#   keys         : torch.Tensor [B, N, C] - Token keys.
#   prune_mode   : Optional[str] - pruning mode; if None, no pruning is applied.
#   sim_threshold: float (default 0.75) - cosine similarity threshold.
#   K            : int (default 5) - minimum valid neighbors.
#   gamma        : float (default 10) - scaling factor for redundancy.
#   class_token  : bool (default False) - flag indicating presence of a class token.
#   distill_token: bool (default False) - flag indicating presence of a distillation token.

def saint_drop(keys, prune_mode, sim_threshold=0.75, K=5, gamma=10,
                              class_token=False, distill_token=False):
    # Return identity if pruning is disabled.
    if prune_mode is None:
        return lambda x: x

    # Exclude protected tokens (e.g., class/distill tokens).
    protected = int(class_token) + int(distill_token)
    if protected:
        keys = keys[:, protected:]

    # Normalize embeddings.
    keys = keys / keys.norm(dim=-1, keepdim=True)
    
    # Bipartite Graph Construction.
    a, b = keys[..., ::2, :], keys[..., 1::2, :]
    scores = a @ b.transpose(-1, -2)
    
    # Measure node degrees and do batch-level voting to calculate r.
    valid   = scores >= sim_threshold
    counts  = valid.sum(dim=-1)
    r = int((counts >= K).float().mean().item())
    if r <= 0:
        return lambda x: x

    # Compute redundancy scores.
    candidate   = (scores * valid).sum(dim=-1) / counts.clamp(min=1)
    candidate   = counts * torch.exp(gamma * (candidate - sim_threshold))
    alternative = scores.mean(dim=-1)
    final_scores= torch.where(counts > 0, candidate, alternative)
    
    # Determine indices for token dropping.
    sorted_idx = final_scores.argsort(dim=-1, descending=True).unsqueeze(-1)
    drop_idx   = sorted_idx[..., r:, :]

    # Define the drop function.
    def drop_func(x):
        if protected:
            tokens = x[:, :protected]
            x = x[:, protected:]
        src, dst = x[..., ::2, :], x[..., 1::2, :]
        B, N1, C = src.shape
        kept = src.gather(dim=-2, index=drop_idx.expand(B, N1-r, C))
        return torch.cat([tokens, kept, dst], dim=1) if protected else torch.cat([kept, dst], dim=1)
    
    return drop_func
\end{lstlisting}
\noindent\rule{\textwidth}{0.4pt}
\end{minipage}
\end{center}
\clearpage

\section{More Ablations on Design Choices}
\label{appendix:supp_ablations}

In this section, we investigate two critical hyperparameters in our method: the similarity threshold, \(\tau\), and the transformer layers where pruning is applied. We evaluate the accuracy-throughput trade-off on two models, ViT-B/16 and ViT-L/16 MAE. By default, \(\tau=0.75\) and pruning is applied to the first half of the transformer layers.

\subsection{Ablating Similarity Threshold}
\autoref{tab:supp_ablate_threshold} reports the results for  $\tau \in [0.7, 0.8]$. As expected, lower \(\tau\) values yield higher throughput at the cost of accuracy, whereas higher \(\tau\) values maintain better accuracy with reduced throughput. Notably, different architectures show different tradeoff behaviour for given thresholds.

\begin{table}[t]
\centering
\caption{Ablation of the similarity threshold \(\tau\) on ViT-B/16 and ViT-L/16 MAE. Lower \(\tau\) increases throughput at the expense of accuracy, and vice versa.}
\label{tab:supp_ablate_threshold}
\resizebox{\columnwidth}{!}{ 
\begin{tabular}{lcc@{\hspace{1.5cm}}cc}
\toprule
 & \multicolumn{2}{c}{\textbf{ViT-B/16}} & \multicolumn{2}{c}{\textbf{ViT-L/16 MAE}} \\
\textbf{Threshold} & \textbf{Acc (\%)} & \textbf{Img/s} & \textbf{Acc (\%)} & \textbf{Img/s} \\
\midrule
\rowcolor{gray!20} 1 (no prune) & 81.0    & 377.1     & 85.96    & 114.4    \\
0.80        & 80.93 & 462.0  & 85.40 & 186.1 \\
0.79        & 80.77 & 469.5  & 85.27 & 191.8 \\
0.78        & 80.66 & 479.0  & 85.19 & 202.6 \\
0.77        & 80.61 & 497.0  & 85.04 & 211.67\\
0.76        & 80.47 & 509.0  & 84.88 & 223.13\\
0.75        & 80.31 & 535.0  & 84.74 & 238.6 \\
0.74        & 80.13 & 554.4  & 84.57 & 242.1 \\
0.73        & 80.05 & 573.0  & 84.43 & 251.2 \\
0.72        & 79.73 & 598.2  & 84.22 & 267.3 \\
0.71        & 79.44 & 630.3  & 84.02 & 274.5 \\
0.70        & 79.01 & 653.0  & 83.94 & 284.33 \\
\bottomrule
\end{tabular}
}
\end{table}

\begin{figure}[t]
  \centering
  \includegraphics[width=\linewidth]{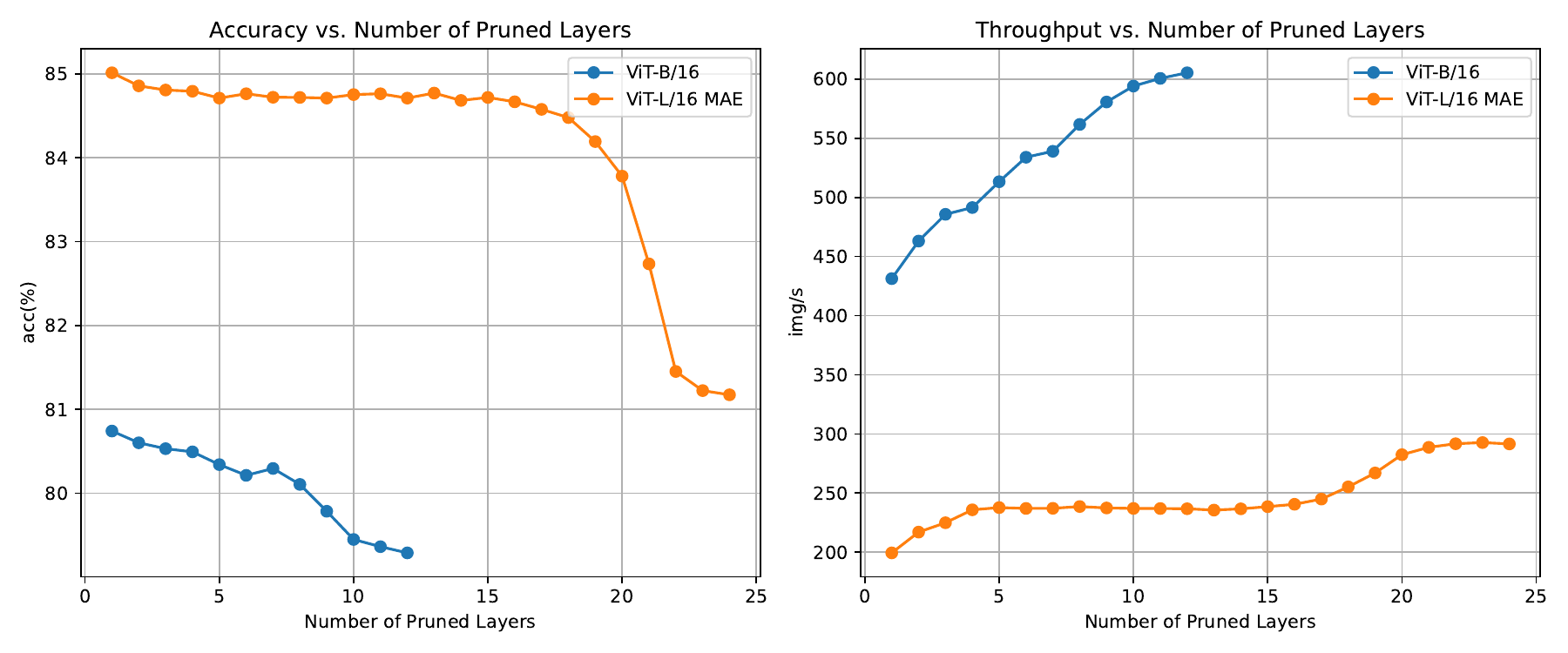}
  \caption{Layer Number Analysis. This figure shows the impact of progressively increasing the number of pruned layers from the beginning of the network. The results indicate that moderate pruning significantly boosts throughput with only a slight reduction in accuracy.}
  \label{fig:layer_number_analysis}
\end{figure}

\begin{figure}[t]
  \centering
  \includegraphics[width=\linewidth]{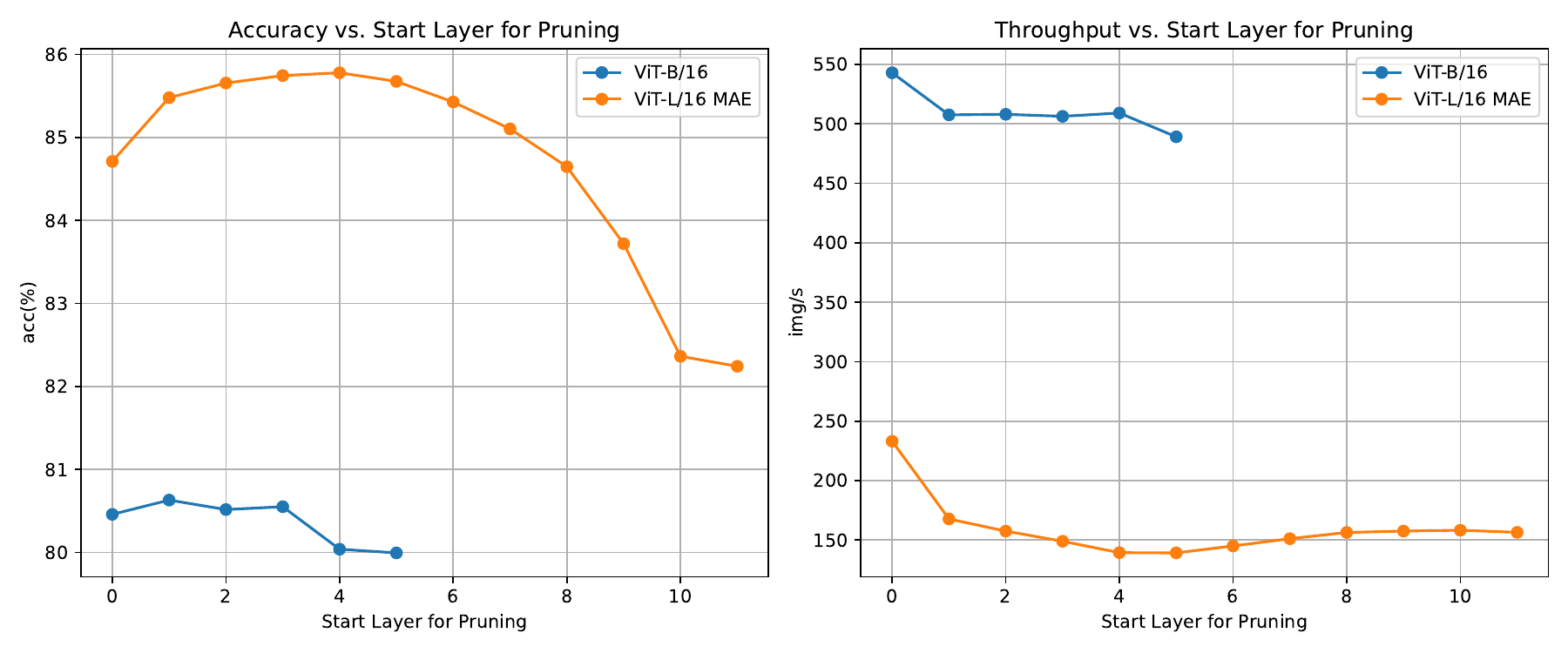}
  \caption{Start Layer Analysis. Here, the effect of shifting the starting layer for pruning in a constant fraction of the network (half) layers is examined. Early stage pruning is shown to yield considerable throughput gains while preserving high accuracy.}
  \label{fig:saint_layer_analysis}
\end{figure}

\subsection{Ablation Study on Pruning Layers in ViTs}
To investigate the influence of pruning layers on the accuracy–throughput tradeoff, we conducted two ablation studies. \autoref{fig:layer_number_analysis} presents the results obtained by progressively increasing the number of pruned layers from the beginning of the network. The data reveal that moderate pruning leads to substantial throughput improvements with only a marginal drop in accuracy. In contrast, \autoref{fig:saint_layer_analysis} explores the effect of varying the starting layer for pruning a fixed portion of the network. The findings suggest that early stage pruning is particularly effective, as it substantially enhances throughput while maintaining robust accuracy. These insights underscore the importance of carefully selecting the pruning layers to optimize overall network performance.

\subsection{Ablation Study on Pruning Layers in VLMs}
\autoref{tab:ablation_llm_only} presents the impact of different pruning configurations in the LLM-Only approach. The table evaluates various layer selections and threshold values ($\tau$) across multiple datasets, analyzing their effects on performance and latency. By adjusting the number and range of layers, as well as modifying $\tau$, we observe how these parameters influence the model's efficiency and accuracy. This ablation study provides insights into the trade-offs between computational cost and task-specific performance, helping to optimize the pruning strategy for improved results.

\begin{table}[t]
    \centering
    \renewcommand{\arraystretch}{1.2} 
    \setlength{\tabcolsep}{2pt} 
    \caption{Ablation of different layers and different $\tau$ on the LLM-Only approach.}
    \footnotesize
    \label{tab:ablation_llm_only}
    \begin{tabular}{lccccc}
        \toprule
        \textbf{Dataset} & \textbf{LLaVA} & \textbf{Layers} & $\boldsymbol{\tau}$ & \textbf{Perf.} & \textbf{Lat.} \\
        \midrule
        MME  & 7B  & $\{2, ..., 31\}$       & 0.95  & 1505  & 159 \\
        MME  & 7B  & $\{2, ..., 8\}$        & 0.88  & 1491  & 110 \\
        MME  & 7B  & $\{6, ..., 13\}$       & 0.86  & 1504  & 152 \\
        POPE & 7B & $\{6, ..., 8\}$        & 0.85  & 85.30 & 133 \\
        POPE & 7B & $\{7, 9, 11, 13, 15\}$ & 0.80  & 86.28 & 151 \\
        GQA  & 7B & $\{7, 9, 11, 13\}$     & 0.80  & 61.31 & 165 \\
        MME    & 13B   & $\{4\}$ & 0.5     & 1515     & 224   \\
        GQA    & 13B   & $\{4\}$ & 0.5     & 62.82     & 254   \\
        POPE    & 13B   & $\{8, ..., 16\}$ & 0.86     & 87.23     & 262   \\
        MME    & 13B   & $\{2, 3\}$ & 0.92     & 1523     & 239   \\
        \bottomrule
    \end{tabular}
\end{table}

\autoref{tab:vit_only_comparison} presents the impact of different pruning configurations in the ViT-Only approach. The table evaluates various layer selections, values of K, and remaining token counts across multiple datasets, analyzing their effects on performance and latency. By adjusting the number of layers and the pruning granularity through K, we examine how these parameters influence the model's efficiency and accuracy. This ablation study highlights the trade-offs between computational cost and task-specific performance, aiding in the optimization of pruning strategies for improved results.

\begin{table}[t]
\centering
\renewcommand{\arraystretch}{1.2} 
\setlength{\tabcolsep}{4pt} 
\footnotesize 
\caption{Ablation of different configurations on performance and latency for ViT-Only approach.}
\label{tab:vit_only_comparison}
\begin{tabular}{lcccccc}
\toprule
\textbf{Dataset} & \textbf{LLaVA} & \textbf{Layer} & \textbf{K} & \textbf{Tokens} & \textbf{Perf.} & \textbf{Lat.} \\
\midrule
MME  & 7B  & 23 & 16 & 377 & 1519  & 287  \\
MME  & 7B  & 23 & 16 & 192 & 1456  & 135  \\
MME  & 7B  & 23 & 16 & 64  & 1287  & 120  \\
MME  & 7B  & 20 & 32 & 377 & 1491  & 165  \\
MME  & 7B  & 20 & 32 & 192 & 1451  & 131  \\
MME  & 7B  & 20 & 32 & 92  & 1332  & 125  \\
POPE & 7B  & 20 & 8  & 192 & 87.01 & 133  \\
POPE & 7B  & 20 & 8  & 92  & 85.81 & 114  \\
POPE & 7B  & 21 & 8  & 288 & 86.70 & 147  \\
POPE & 7B  & 22 & 8  & 288 & 86.69 & 120  \\
GQA  & 7B  & 20 & 4  & 144 & 59.03 & 142  \\
GQA  & 7B  & 20 & 4  & 92  & 57.62 & 179  \\
POPE & 13B & 23 & 8  & 288 & 85.51 & 216  \\
MME  & 13B & 23 & 16 & 64  & 1319  & 168  \\
MME  & 13B & 23 & 16 & 144 & 1416  & 179  \\
MME  & 13B & 23 & 16 & 288 & 1511  & 218  \\
\bottomrule
\end{tabular}
\end{table}

\section{FLOPs Analysis}
\label{appendix:supp_flops}

Many works in token pruning focus on reducing GFLOPs as a proxy for computational efficiency. However, a lower GFLOP count does not always translate into better overall performance. \autoref{tab:gflops} compares the accuracy, image throughput (img/s), and GFLOPs for two models when throughput is roughly doubled relative to the baseline. Although our method SAINT exhibits a GFLOPs count slightly above some competitors (e.g., ToMe and PPT), it achieves a substantially higher accuracy. This finding reinforces the advantage of pruning tokens early in the network and optimizing for actual throughput and efficiency rather than FLOPs.

\begin{table}[t]
  \centering
  \caption{Performance and computational cost comparison for ViT-H/14 MAE and ViT-L/14 MAE models at roughly doubled throughput. Although SAINT shows a modest GFLOPs reduction relative to the baseline, its superior accuracy demonstrates the benefits of early token pruning over methods that aggressively minimize GFLOPs.}
  \label{tab:gflops}
  \resizebox{\columnwidth}{!}{ 
  \begin{tabular}{llccc}
    \toprule
    \textbf{Model} & \textbf{Prune Method} & \textbf{Acc (\%)} & \textbf{img/s} & \textbf{GFLOPs} \\
    \midrule
    \multirow{5}{*}{ViT-H/14 MAE - 224px} 
      & Baseline         & 86.88 & 43.5  & 161.9  \\
      & ToMe (r=8)       & 85.45 & 86.8  & 82.5   \\
      & PPT (r=30)       & 85.26 & 82.8  & 82.6   \\
      & ToFu (r=8)       & 83.30 & 88.9  & 100.8  \\
      & SAINT ($\tau=0.8$)  & 86.21 & 86.4  & 89.0   \\
    \midrule
    \multirow{5}{*}{ViT-L/14 MAE - 224px}  
      & Baseline         & 85.96 & 114.5 & 59.7   \\
      & ToMe (r=8)       & 84.21 & 225.7 & 31.0   \\
      & PPT (r=40)       & 83.74 & 229.8 & 25.9   \\
      & ToFu (r=8)       & 82.11 & 225.4 & 37.6   \\
      & SAINT ($\tau=0.75$) & 84.74 & 238.6 & 31.5   \\
    \bottomrule
  \end{tabular}
  }
\end{table}